\documentclass[conference]{IEEEtran}
\IEEEoverridecommandlockouts
\usepackage{setspace}
\usepackage{cite}
\usepackage{amsmath,amssymb,amsfonts,mathtools,bm}
\usepackage{amsthm}
\usepackage{algorithm}
\usepackage{graphicx}
\usepackage{textcomp}
\usepackage{xcolor,float}
\usepackage{tabularx}
\usepackage{textcomp}
\usepackage{diagbox}
\usepackage{svg}
\usepackage{booktabs}
\usepackage{subfigure}
\usepackage{hyperref}

\newtheorem{problem}{Problem}
\usepackage{algpseudocode}
\usepackage{graphicx}
\usepackage{makecell}
\usepackage{amsthm}
\usepackage{caption}
\usepackage{sidecap}
\usepackage{microtype}
\usepackage{booktabs} 
\usepackage{multirow}
\usepackage{float}
\usepackage{adjustbox} 
\usepackage{amsmath}
\usepackage{amssymb}
\usepackage{colortbl}
\usepackage{makecell}
\usepackage{float}
\usepackage{url}
\usepackage{balance}
\usepackage{bbding}
\usepackage{hyperref}
\usepackage{graphicx}
\usepackage{sidecap}
\usepackage{microtype}
\usepackage{graphicx}
\usepackage{subfigure}
\usepackage{booktabs} 
\usepackage{multirow}
\usepackage{array}
\usepackage{subcaption}
\usepackage{graphicx}
\usepackage{makecell}
\usepackage{amsmath}
\usepackage{amssymb}

\usepackage{tikz}

\usepackage{changes}
\usepackage{ifthen}
\newboolean{showchanges}
\setboolean{showchanges}{false}  % True for blue

\newcommand{\add}[1]{%
    \ifthenelse{\boolean{showchanges}}%
        {\textcolor{blue}{#1}}
        {#1\relax}
}

\definecolor{lime}{HTML}{A6CE39}
\DeclareRobustCommand{\orcidicon}{%
    \begin{tikzpicture}
    \draw[lime, fill=lime] (0,0) 
    circle [radius=0.16] 
    node[white] {{\fontfamily{qag}\selectfont \tiny ID}};    \draw[white, fill=white] (-0.0625,0.095) 
    circle [radius=0.007];    \end{tikzpicture}
    \hspace{-2mm}}
\foreach \x in {A, ..., Z}{%
    \expandafter\xdef\csname orcid\x\endcsname{\noexpand\href{https://orcid.org/\csname orcidauthor\x\endcsname}{\noexpand\orcidicon}}
    }
\allowdisplaybreaks
\renewcommand{\baselinestretch}{1}

\begin{document}
% \doublespacing

\title{iRadioDiff: Physics-Informed Diffusion Model for Indoor Radio Map Construction and Localization}
\author{\IEEEauthorblockN{Xiucheng Wang\IEEEauthorrefmark{1},
Tingwei Yuan\IEEEauthorrefmark{2},
Yang Cao\IEEEauthorrefmark{3}
Nan Cheng\IEEEauthorrefmark{1},
Ruijin Sun\IEEEauthorrefmark{1},
Weihua Zhuang\IEEEauthorrefmark{4}}
\IEEEauthorblockA{
\IEEEauthorrefmark{1}State Key Laboratory of ISN and School of Telecommunications Engineering, Xidian University, Xi'an, 710071, China\\
\IEEEauthorrefmark{2}School of Artificial Intelligence, Xidian University, Xi'an, 710071, China\\
\IEEEauthorrefmark{3}School of Information Science and Technology, Southwest Jiaotong University, Chengdu, China\\
\IEEEauthorrefmark{3} Department of Electrical and Computer Engineering, University of Waterloo, Waterloo, ON N2L 3G1, Canada\\
Email: \{xcwang\_1, 23009201393\}@stu.xidian.edu.cn, cyang9502@gmail.com, dr.nan.cheng@ieee.org,\\ sunruijin@xidian.edu.cn, wzhuang@uwaterloo.ca }
}
   % , wzhuang@uwaterloo.ca 
    \maketitle

\IEEEdisplaynontitleabstractindextext

\IEEEpeerreviewmaketitle

\begin{abstract}
Radio maps (RMs) serve as environment-aware electromagnetic (EM) representations that connect scenario geometry and material properties to the spatial distribution of signal strength, enabling localization without costly in-situ measurements. However, constructing high-fidelity indoor RMs remains challenging due to the prohibitive latency of EM solvers and the limitations of learning-based methods, which often rely on sparse measurements or assumptions of homogeneous material, which are misaligned with the heterogeneous and multipath-rich nature of indoor environments. To overcome these challenges, we propose iRadioDiff, a sampling-free diffusion-based framework for indoor RM construction. iRadioDiff is conditioned on access point (AP) positions, and physics-informed prompt encoded by material reflection and transmission coefficients. It further incorporates multipath-critical priors, including diffraction points, strong transmission boundaries, and line-of-sight (LoS) contours, to guide the generative process via conditional channels and boundary-weighted objectives. This design enables accurate modeling of nonstationary field discontinuities and efficient construction of physically consistent RMs. Experiments demonstrate that iRadioDiff achieves state-of-the-art performance in indoor RM construction and received signal strength based indoor localization, which offers effective generalization across layouts and material configurations. Code is available at \url{https://github.com/UNIC-Lab/iRadioDiff}.
\end{abstract}

\section{Introduction}
Indoor localization plays a vital role in smart buildings, industrial automation, emergency response, and immersive applications, with increasing demands for meter-to-submeter accuracy and scalable, low-latency deployment \cite{zafari2019survey,6g}. However, existing approaches face fundamental limitations. Ranging-based methods, relying on received signal strength information (RSSI) or time of arrival (ToA) with statistical distance–loss models, often suffer from significant range-estimation mismatch in complex indoor environments due to multipath, penetration, and diffraction effects \cite{shamsfakhr2022indoor}. Fingerprint-based methods, while more robust to propagation variability, require extensive offline surveys and frequent updates, incurring high cost and reduced scalability \cite{wang2016csi}. Alternatively, the radio map (RM) offers an environment-aware electromagnetic representation that links scenario geometry and materials to the spatial distribution of path loss or RSSI \cite{zeng2021toward}. RMs provide physically consistent priors for localization and, when generated via electromagnetic (EM) simulation or neural network (NN)-based methods, can significantly improve the indoor localization performance \cite{zeng2024tutorial}.

Constructing high-fidelity indoor RMs remains challenging. Full-wave EM solvers, while accurate, are limited to small-scale domains due to prohibitive memory and computational demands, making them impractical for room-scale deployments \cite{jones2013theory}. Approximate methods such as ray tracing offer partial acceleration, but still incur minute-level latency under complex multipath, rendering them unsuitable for frequent updates triggered by door or furniture changes \cite{deschamps1972ray}. Neural models enable faster inference, but most NN-based indoor RM methods rely on costly sparse measurements, limiting their use in zero-measurement settings \cite{zhang2023rme}. Furthermore, existing sampling-free RM construction approaches are typically designed for outdoor environments, assuming homogeneous materials \cite{levie2021radiounet,wang2024radiodiff}. Such assumptions are generally invalid for indoor environments, where heterogeneous materials and complex multipath propagation dominate, leading to a mismatch between model design and practical deployment conditions. These factors lead to inaccuracies near boundaries and RSSI discontinuities, where purely data-driven models often fail to capture material-sensitive propagation behavior. This motivates a new approach capable of fast, measurement-free RM generation that incorporates material and boundary awareness while maintaining physical consistency for real-time indoor localization.

\begin{figure*}[ht]
    \centering
    \includegraphics[width=0.8\linewidth]{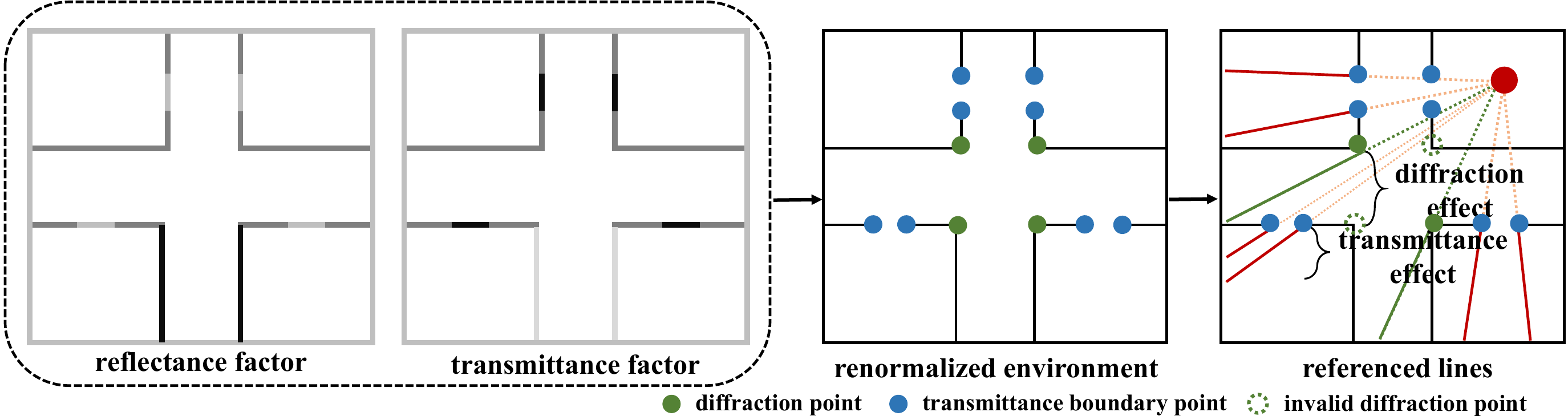}
    \caption{Illustration of the multipath boundary extraction. The darker boundary means a higher reflectance and transmittance factor for both figures of reflectance and transmittance.}
    \label{fig-boundary}
    \vspace{-15pt}
\end{figure*}

To address the above challenges, motivated by the success of generative AI in outdoor RM construction \cite{wang2024radiodiff}, we propose iRadioDiff, which brings diffusion models (DMs) to sampling-free indoor RM generation. To overcome the implicit homogeneity assumptions prevalent in outdoor pipelines, we encode indoor material reflection and transmission coefficients as physics-aware prompts to the DM. In addition, leveraging the interplay between EM parameters and scenario geometry, we extract multipath-critical structures, namely diffraction points and strong transmission boundaries, as is shown in Fig.~\ref{fig-boundary}; together with line-of-sight (LoS) masks, they define an RSSI-discontinuity prior. During training and inference, the prior is injected as conditional channels/masks and enforced via boundary-weighted learning signals, thereby constraining the score network to respect material and boundary laws of EM propagation. The main contributions of this paper are summarized as follows.
\begin{enumerate}
    \item To the best of our knowledge, we first employ diffusion models for zero-measurement indoor RM construction. Using access point (AP) locations with material EM parameters as inputs, our approach removes the implicit homogeneous-medium assumption common in outdoor pipelines, achieves robust generalization across material heterogeneity and layout changes, and supports rapid re-generation by simply updating the physics prompts.
    \item We encode reflection and transmission coefficients as physics prompts to the diffusion model and explicitly extract diffraction points and strong-transmission boundaries. Combined with LoS masks, they define an RSSI-discontinuity prior that is injected via conditional channels and enforced with boundary-weighted learning signals, markedly improving the representation of strongly non-stationary structures while preserving physical consistency.
    \item Experimental results demonstrate that iRadioDiff achieves state-of-the-art (SOTA) performance in indoor RM construction and localization accuracy using RSSI-only information. Ablation studies confirm the validity of the proposed physics-driven information for effective indoor RM construction.
\end{enumerate}

\section{Preliminary and System Model}
\subsection{Diffusion Model}
DMs have emerged as powerful generative frameworks capable of modeling complex data distributions through iterative denoising. In score-based diffusion models, the forward process is formulated as a stochastic differential equation (SDE) $d\bm{x} = f(\bm{x}, t)dt + g(t)d\bm{w}$, where $f(\bm{x}, t)$ and $g(t)$ define the drift and diffusion coefficients, respectively. As $t$ increases, data distribution $p_t(\bm{x})$ converges to an isotropic Gaussian. Sample generation is performed by integrating the reverse-time SDE as $d\bm{x} = \left[f(\bm{x}, t) - g^2(t)\nabla_{\bm{x}}\log p_t(\bm{x})\right]dt + g(t)d\bar{\bm{w}}$, with score function $\nabla_{\bm{x}}\log p_t(\bm{x})$ approximated by a neural network, $s_\theta(\bm{x}, t)$. For deterministic sampling, an equivalent probability flow ODE can be used, replacing stochasticity with a drift-only formulation.

To improve generative stability and sampling quality, particularly in RM construction, decoupled diffusion models (DDMs) have been introduced by the SOTA outdoor RM construction method RadioDiff \cite{wang2024radiodiff}.  The DDM stabilizes generation by separating deterministic signal contraction from stochastic noise injection. Instead of directly noising the data, DDMs first drive $\bm{x}_0$ along a deterministic flow field $\bm{f}_t$ toward a zero baseline, then add Gaussian noise. Under a canonical parametrization with unit diffusion, the forward marginal admits a closed form \cite{huang2024simultaneous}
\begin{align}
q(\bm{x}_t|\bm{x}_0) = \mathcal{N}\left(\bm{x}_0 + \int_0^t \bm{f}_\tau \, d\tau,\; t\,\bm{I}\right),\label{ddm-forward}
\end{align}
which makes the attenuation path explicit and renders the noise variance purely time–dependent. The reverse-time update is likewise Gaussian and enables efficient sampling with a single-step closed form \cite{huang2024simultaneous}
\begin{align}
q\left(\bm{z}_{t-\Delta t} \mid \bm{z}_t, \bm{z}_0\right)  &=\mathcal{N}\left(\bm{z}_{t} +\int_t^{t-\Delta t} \bm{f}_t \mathrm{~d} t\right. \notag\\
& \left.\qquad\qquad-\frac{\Delta t}{\sqrt{t}} \bm{\epsilon}_{t}, \frac{\Delta t(t-\Delta t)}{t} \bm{I}\right),
\end{align}
where $\bm{\epsilon}_{t}$ is a stander Gaussian random variable. By decoupling geometry-aware contraction ($\bm{f}_t$) from stochastic perturbation, DDMs yield better conditioned training objectives and smoother reverse trajectory properties.

\begin{figure*}[ht]
    \centering
    \includegraphics[width=0.8\linewidth]{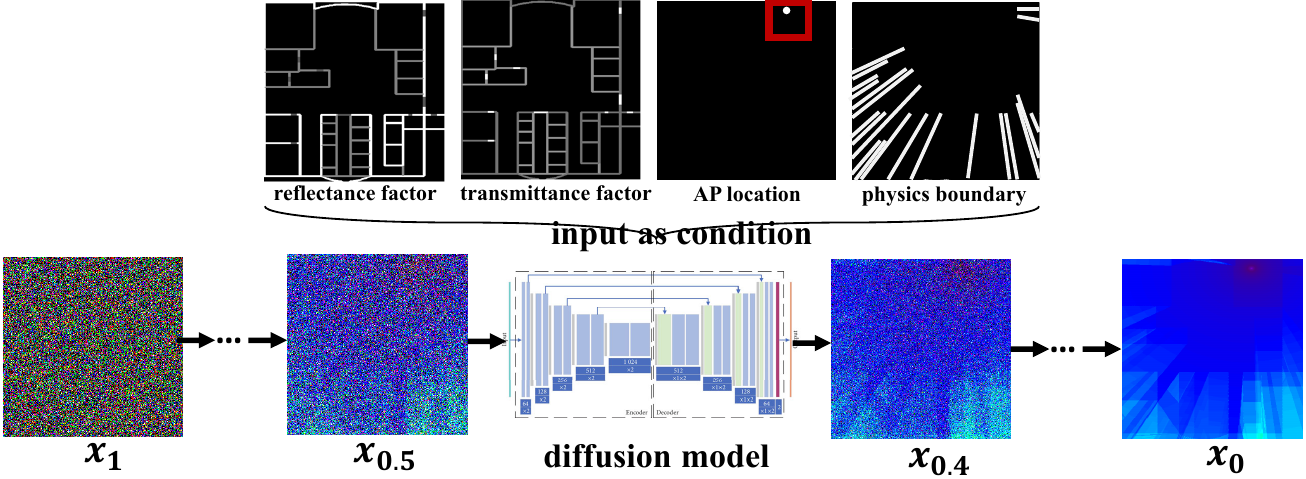}
    \caption{Illustration of the iRadioDiff framework.}
    \vspace{-12pt}
    \label{fig-system}
\end{figure*}

\subsection{System Model and Problem Formulation}
Consider a two-dimensional indoor region $\Omega \subset \mathbb{R}^2$, discretized into uniform grid $\mathcal{G}$ with spatial resolution $\Delta$. The environment contains a single AP located at known position $\bm{s}$. To capture material heterogeneity and boundary effects, we define two co-registered spatial fields $H_r(\bm{x})$ and $H_t(\bm{x})$, representing the effective reflection and transmission coefficients at each location $\bm{x} \in \mathcal{G}$, respectively. The ground-truth RM is denoted by $\bm{P}^\ast \in \mathbb{R}^{|\mathcal{G}|}$, where $P^\ast(\bm{x})$ specifies the RSSI at location $\bm{x}$ from the AP. We operate under a zero-measurement regime, where the RM is generated solely from prior knowledge of the environment and AP configuration, without access to any in-situ RSSI measurements. Notably, prior theoretical studies have demonstrated that highly accurate indoor RSSI information enables precise positioning using traditional methods such as k-nearest neighbors (KNN) or likelihood-based localization. Therefore, in the absence of direct signal measurements, our objective is equivalent to constructing a high-fidelity indoor RM, so as to indirectly enable accurate RSSI-based localization. To this end, we define the input condition as $\mathcal{C} = \{\bm{s}, H_r, H_t\}$. A neural network, $F_\theta$, is trained to map $\mathcal{C}$ to an estimated RM $\widehat{\bm{P}} = F_\theta(\mathcal{C})$, using dataset $\mathcal{D} = \{ (\mathcal{C}^{(n)}, \bm{P}^{\ast(n)}) \}_{n=1}^N$ generated from simulation or curated reference data. The objective is given by
\begin{problem}
    \begin{align}
        &\min_{\theta} \ \frac{1}{N} \sum_{n=1}^{N} \big\| F_\theta(\mathcal{C}^{(n)}) - \bm{P}^{\ast(n)} \big\|_2^2.\label{obj}
    \end{align}
\end{problem}

\section{Physics Informed DM for Indoor RM Generation}

\subsection{Physics-Informed Environment Feature Extraction}
To enhance the physical fidelity of RM generation for indoor localization, we introduce a geometric preprocessing mechanism that encodes field-discontinuity priors into the generative process. This design is motivated by a key observation: abrupt changes in RSSI commonly arise near diffraction points, such as sharp corners, wedges, and strong transmission boundaries, such as doors and windows, where EM wave behavior violates the local continuity assumption underpinning convolutional neural networks. These discontinuities manifest as localized non-smooth variations in the RM, impeding conventional learning and interpolation techniques. To address this, we extract a set of diffraction point candidates, $\mathcal{C} = {c_k}$, from the environment geometry. A point is defined as a valid corner candidate if it satisfies the following geometric rule: its own reflection coefficient is non-zero, which cannot be air, and among its four direct neighbors as up/down/left/right, at least one has a non-zero reflection coefficient while at least two others are zero. This geometric configuration corresponds to a convex discontinuity in reflectance, a necessary condition for wedge diffraction to occur. In parallel, we extract strong transmission boundary points $\mathcal{B}_t = {b_j}$ corresponding to high-transparency regions such as doors, windows, or movable partitions, derived from transmission coefficient maps. Given the position $\bm{s}$ of an AP, we apply a two-stage geometric filtering process. First, ray-connected boundary extraction is conducted by linking spatially adjacent points in $\mathcal{C} \cup \mathcal{B}_t$ along walls or boundary surfaces. We remove all radial segments that directly connect $\bm{s}$ to any $c_k$ or $b_j$, thus avoiding trivial source–scatterer paths and retaining only plausible field-discontinuity contours. Second, directional culling is performed for diffraction points. Based on geometrical optics, if the AP and the outward-facing normal vector of the obstacle hosting $c_k$ lie in the same directional quadrant, then $c_k$ is considered invalid, as it cannot form an effective shadowed wedge required for diffraction. This implements the physical ``no-shadow, no-diffraction" rule. Moreover, we further identify a class of non-effective diffraction points: even if $c_k$ satisfies the wedge configuration and lies on a wall between LoS and NLoS zones, it may fail to induce a perceptible RSSI transition if it resides in the same room as the AP. In such cases, the corresponding wall already attenuates both LoS and NLoS propagation paths equally, making the corner geometrically redundant in terms of field discontinuity. These points are also removed. To complement the diffraction-based discontinuity detection, we incorporate LoS boundary extraction as an additional conditioning prior. Using an efficient rotational scanning method from the AP location $\bm{s}$, we compute the LoS field across the domain and extract the LoS–NLoS transition contour. This boundary typically manifests as a smooth shadow front behind structural occluders and serves as a complementary prior channel to indicate the signal intensity envelope. The final prior representation is denoted as $\bm{c} = \{\bm{s}, H_r, H_t, \Gamma(\bm{s}, \widetilde{\mathcal{C}}, \mathcal{B}_t), \mathrm{LoS}(\bm{s})\}$, where $\Gamma$ encodes the pruned diffraction and transmission-induced contours, and $\mathrm{LoS}(\cdot)$ denotes the extracted LoS mask. By informing the generative model of where sharp field transitions or occlusion-induced variations are likely to occur, we improve its ability to synthesize physically consistent RM samples that preserve edge-aware spatial characteristics critical for downstream localization tasks.

\subsection{Physics Enhanced Diffusion Model}
Building upon the physics-informed priors introduced above, we adopt a continuous-time diffusion probabilistic model to learn the conditional distribution of indoor radio maps. Unlike latent-domain approaches, we operate directly in the image space of $\bm{P}^\ast$ to preserve fine-grained spatial structures essential for downstream localization. The generative backbone is instantiated as a U-Net with temporal input $t \in [0,1]$ and a set of conditioning inputs integrated via cross-attention. Specifically, the conditioning vector comprises (i) AP location $\bm{s}$, encoded as a Gaussian heatmap centered at $\bm{s}$; (ii) reflection coefficient map $H_r(\bm{x})$ and transmission coefficient map $H_t(\bm{x})$; and (iii) the discontinuity-aware priors $\Gamma(\bm{s}, \widetilde{\mathcal{C}}, \mathcal{B}_t)$ and LoS-aware information derived from geometry-based analysis. These multi-channel inputs are collectively fed into the U-Net’s attention layers using a spatial cross-attention mechanism, allowing the network to dynamically adapt its denoising trajectory based on material and topological context. During training, we sample a random timestep $t \sim \mathcal{U}(0,1)$ of the diffusion process in \eqref{ddm-forward} and perturb the ground-truth RM, $\bm{x}_0$, according to the forward stochastic differential equation \eqref{ddm-forward}, where $\bm{f}_t$ denotes the drift vector field. The network is trained to predict both drift $\widehat{\bm{f}}_t$ and added noise $\widehat{\bm{\epsilon}}_t$ at each step, using a mean-squared error loss between the predictions and ground-truth values. The overall objective encourages the model to approximate the score function $\nabla_{\bm{x}_t} \log p(\bm{x}_t|\bm{c})$, facilitating accurate reverse-time denoising. At inference time, we initialize the process with a stander Gaussian sample, $\bm{x}_t \sim \mathcal{N}(0,\bm{I})$, and apply the reverse-time dynamics, conditioned on $\bm{c}$, to iteratively recover denoised signal $\bm{x}_0$, which corresponds to the final synthesized RM. This framework allows the model to explicitly incorporate physical structure and boundary-induced discontinuities into the generative trajectory, enabling it to synthesize RSSI fields that conform more closely to real-world propagation characteristics, especially in cluttered indoor environments where naive interpolation fails.

\section{Experiment Results}
\subsection{Datasets and Implementation Details}
We evaluate our method on a subset of the publicly available Indoor Radio Map Dataset \cite{bakirtzis2025first}, which provides RSSI radio maps generated via ray tracing under various indoor propagation conditions. Ray tracing offers a principled approximation to Maxwell’s equations, yielding physically grounded RSSI distributions that capture key multipath behaviors including reflection, transmission, and diffraction. Among the available settings, we focus exclusively on the 3.5 GHz sub-6G band and ignore differences in antenna radiation patterns to isolate the influence of environment-induced propagation effects. Each environment is discretized at 0.25 m spatial resolution, with propagation paths computed using up to 8 reflections, 10 transmissions, and 2 diffractions. The RSSI values are recorded at receiver height 1.5 m, matching the transmitter height. To validate generalization, we consider two evaluation protocols across 25 distinct indoor layouts.
\begin{itemize}
    \item \textbf{Antenna location generalization (ALG)}: All 25 layouts are included during training, but only 40 out of 50 AP positions per layout are used for training, while the remaining 10 are used for testing—assessing the model’s ability to generalize across AP deployments.
    \item \textbf{Zero-shot layout generalization (ZLG)}: 20 environments, each with 50 APs, are used for training, while the remaining 5 unseen environments, each with 50 APs, are held out for testing. This setting is to evaluate the model’s robustness under previously unseen topological and material configurations.
\end{itemize}
To ensure a fair and comprehensive comparison, we evaluate our proposed method against several representative deep learning-based RM construction models, each reflecting different architectural paradigms. All baseline methods are trained and tested using the same datasets and experimental settings as iRadioDiff to ensure consistent evaluation. The compared methods are as follows.
\begin{itemize}
\item \textbf{RadioUNet}\cite{levie2021radiounet}: A widely used U-Net CNN that maps environmental descriptors to RSSI fields and offers a strong efficiency–accuracy baseline, though its convolutional inductive bias can struggle with abrupt field discontinuities.
\item \textbf{RME-GAN}~\cite{zhang2023rme}: A generative adversarial network (GAN) based RM construction method, which builds the RM only based on the environment and AP information.
\item \textbf{SIP2Net}~\cite{SIP2Net}: A recent SOTA indoor RM predictor built on a U-Net backbone augmented with asymmetric convolutions and atrous spatial pyramid pooling (ASPP).
\end{itemize}

\subsection{Performance Comparison}
\begin{figure*}[ht]
    \centering
    \includegraphics[width=1.0\linewidth]{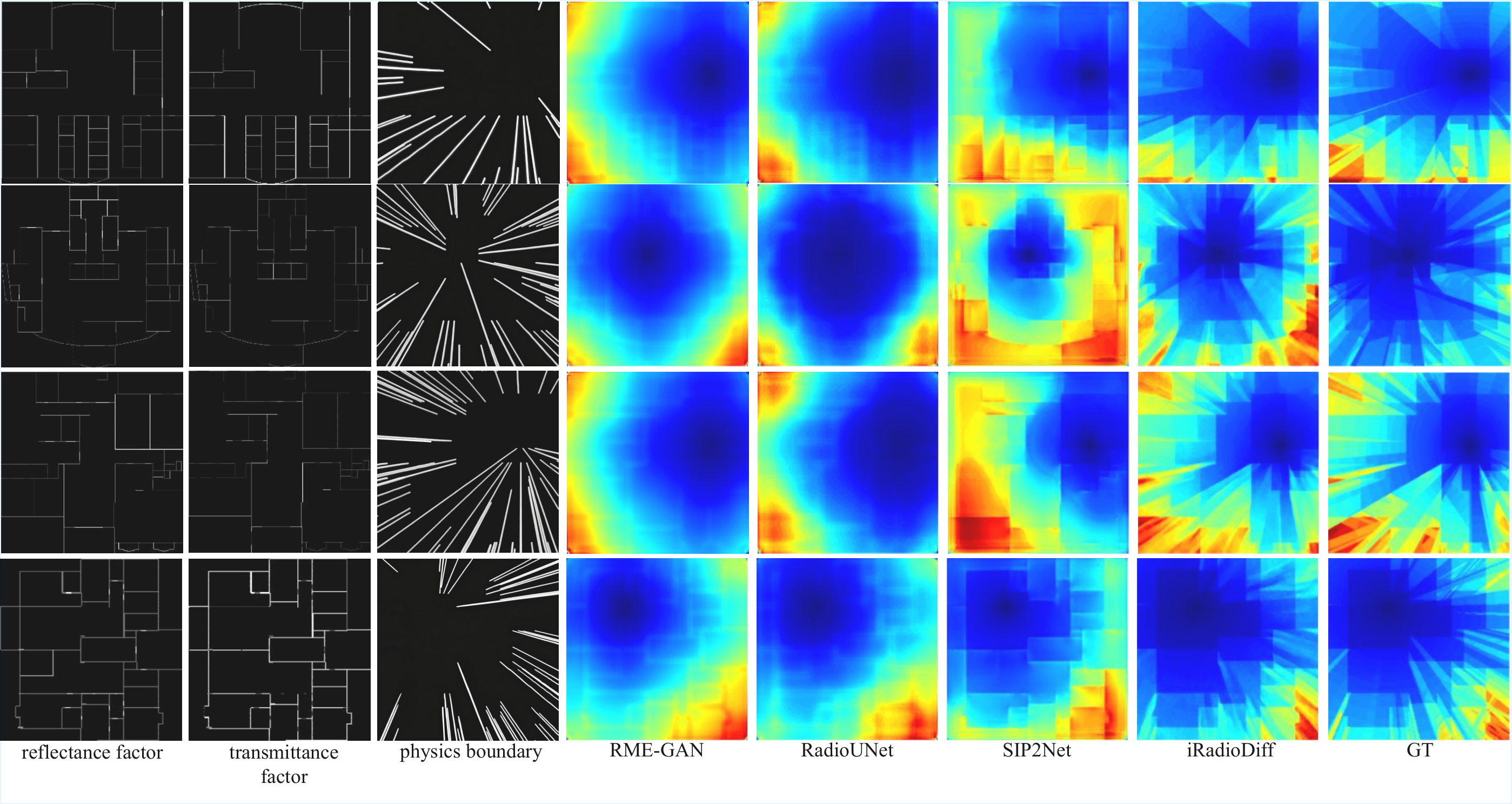}
    \caption{Illustration of the generated RM from different methods in antenna generalization scenarios.}
    \label{fig-antenna-example}
    \vspace{-12pt}
\end{figure*}
\begin{figure*}[ht]
    \centering
    \includegraphics[width=1.0\linewidth]{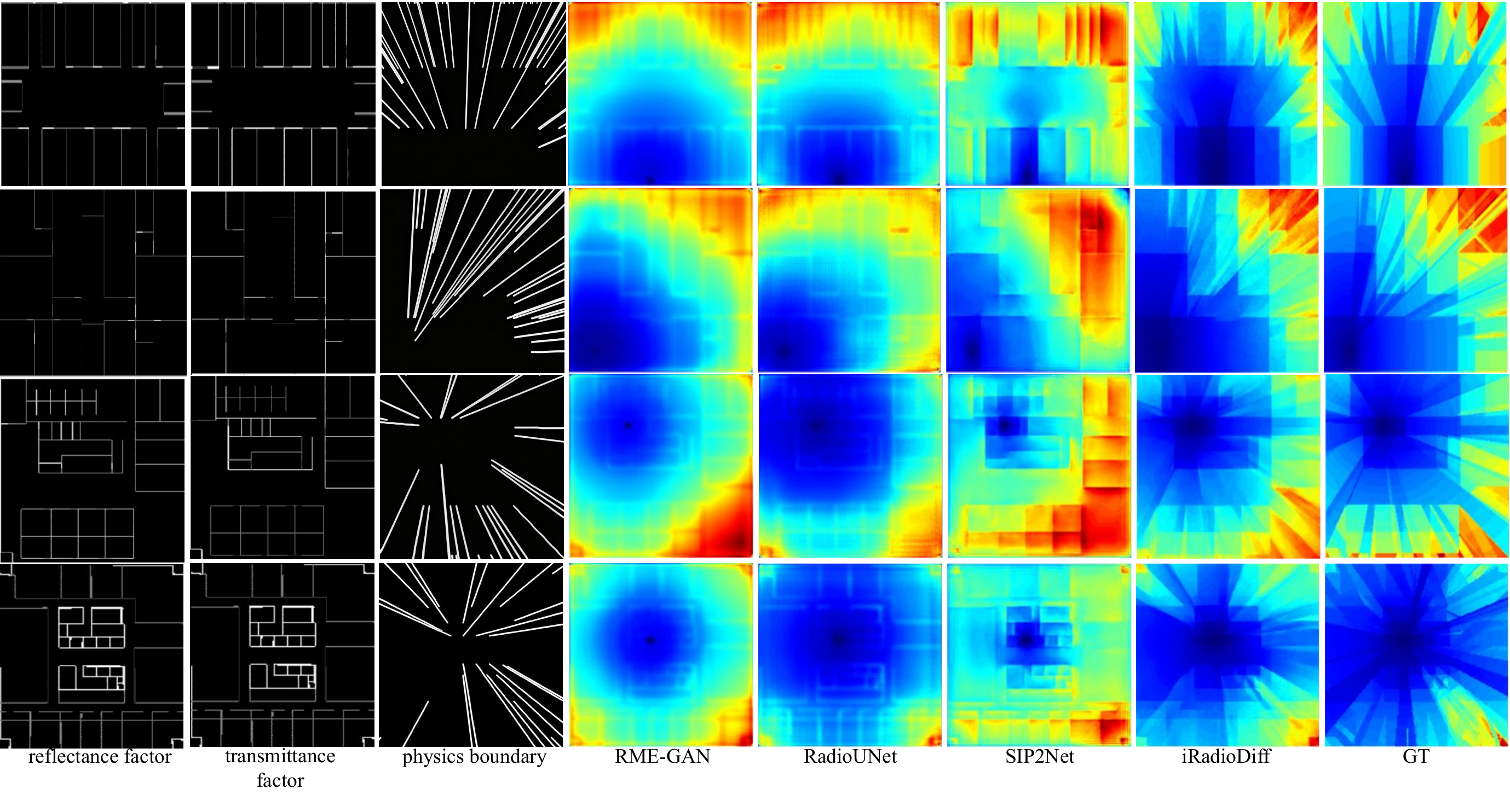}
    \caption{Illustration of the generated RM from different methods in zero-shot layout generalization scenarios.}
    \label{fig-layout-example}
    \vspace{-12pt}
\end{figure*}

\begin{table}[ht]
\captionsetup{font={small}, skip=14pt}
  \centering
  \caption{\textbf{Quantitative Comparison on RM Construction.} Results shown in bold red and with blue underlines indicate the highest and second-highest results, respectively. An upward arrow $\uparrow$ denotes metrics for which larger values indicate better performance; for metrics without this marker, smaller values are preferred.}
  \vspace{-6pt}
\resizebox{\linewidth}{!}{
\begin{tabular}{@{}cc|cccc@{}}
\toprule
\multicolumn{2}{c|}{Methods} & RME-GAN & RadioUNet & SIP2Net & iRadioDiff (Ours) \\ \midrule
 & RMSE & 103.1 & {\color[HTML]{00009B} \underline{9.349}} & 24.55 & {\color[HTML]{9A0000} \textbf{6.357}} \\
 & PSNR $\uparrow$& 7.902 & {\color[HTML]{00009B} \underline{28.92}} & 21.24 & {\color[HTML]{9A0000} \textbf{32.24}} \\
 & LPIPS & 0.5459 & 0.3667 & {\color[HTML]{00009B} \underline{0.3121}} & {\color[HTML]{9A0000} \textbf{0.2742}} \\
\multirow{-4}{*}{ALG} & FID  & 308.1 & 240.12 & {\color[HTML]{00009B} \underline{196.8}} & {\color[HTML]{9A0000} \textbf{145.2}} \\ \midrule
 & RMSE & 93.83 & {\color[HTML]{00009B} \underline{7.868}} & 47.43 & {\color[HTML]{9A0000} \textbf{7.010}} \\
 & PSNR $\uparrow$& 8.723 & {\color[HTML]{00009B} \underline{30.52}} & 15.32 & {\color[HTML]{9A0000} \textbf{31.45}} \\
 & LPIPS & 0.5236 & 0.4188 & {\color[HTML]{9A0000} \textbf{0.3289}} & {\color[HTML]{00009B} \underline{0.3301}} \\
\multirow{-4}{*}{ZLG} & FID & 270.5 & 309.1 & {\color[HTML]{00009B} \underline{197.2}} & {\color[HTML]{9A0000} \textbf{192.6}} \\ \bottomrule
\end{tabular}
}
\vspace{-2mm}
  \label{tab1}
\end{table}
To evaluate the performance of RM generation, we adopt both pixel-wise and perceptual metrics, including root mean square error (RMSE), peak signal-to-noise ratio (PSNR), learned perceptual image patch similarity (LPIPS) \cite{zhang2018unreasonable}, and Fréchet inception distance (FID) \cite{LDM}. RMSE $= \sqrt{\frac{1}{N} \sum_{i=1}^{N} (x_i - \hat{x}i)^2}$ and PSNR $ = 10 \log{10} \frac{MAX^2}{\text{MSE}}$ quantify construction accuracy in terms of absolute and relative error. The LPIPS measures perceptual similarity by comparing deep feature distances from pretrained networks \cite{zhang2018unreasonable}, while FID evaluates the distributional alignment between real and generated samples in the latent space of an Inception model \cite{LDM}. Together, these metrics capture both numerical fidelity and perceptual realism.

Fig.~\ref{fig-antenna-example} and Fig.~\ref{fig-layout-example} provide a visual comparison between the radio maps generated by the proposed iRadioDiff framework and those produced by representative baseline methods, alongside the GT maps obtained via ray tracing. As shown, iRadioDiff consistently reconstructs fine-grained structural details and preserves spatial variations that arise from complex electromagnetic propagation phenomena, such as multipath reflections, diffraction, and partial transmission. In particular, it captures abrupt RSSI transitions and localized shadowing effects with higher fidelity, while baseline methods often oversmooth or misrepresent these discontinuities due to limited physical awareness. Quantitative results summarized in Table~\ref{tab1} further validate these observations. Across both the antenna generalization and zero-shot layout generalization tasks, iRadioDiff achieves SOTA performance on two widely adopted pixel-level construction metrics as RMSE and PSNR. These metrics reflect the model’s ability to recover accurate field intensities and preserve local signal dynamics under both seen and unseen configurations. In addition, we evaluate perceptual consistency using LPIPS and FID, which measure human-aligned similarity and distributional divergence, respectively. iRadioDiff also attains leading performance on these metrics in most test cases, indicating its ability to generate physically consistent and visually realistic radio maps that generalize well across diverse indoor scenarios.

\begin{table}[t]
\centering
\caption{\textbf{Quantitative Comparison of Localization.}}
\vspace{-4pt}
\captionsetup{font={small}, skip=8pt}
\resizebox{0.9\linewidth}{!}{
\begin{tabular}{@{}c|cccc@{}}
\toprule
Methods & RME-GAN & RadioUNet & SIP2Net & iRadioDiff (Ours) \\ \midrule
ALG & 29.60 & {\color[HTML]{00009B} \underline{12.09}} & 21.41 &{\color[HTML]{9A0000} \textbf{7.860}}  \\ \midrule
ZLG & 19.33 & {\color[HTML]{00009B} \underline{12.16}} & 25.40 &{\color[HTML]{9A0000} \textbf{8.530}}   \\ \bottomrule
\end{tabular}
}
\vspace{-12pt}
\label{tab2}
\end{table}
To further evaluate the utility of the constructed RMs in real-world applications, we assess localization performance using a standard KNN algorithm with $K=5$. For each indoor environment, we randomly select 3,000 test locations and compute the average localization error. As shown in Table~\ref{tab2}, RMs generated by iRadioDiff enable the most accurate localization, achieving the lowest average error across all methods. Notably, iRadioDiff is the only method that consistently achieves sub–10-meter accuracy across test cases, demonstrating its superiority in producing high-fidelity RMs that preserve critical signal variations for precise positioning.

\subsection{Ablation Study}
\begin{table}[t]
\centering
\caption{Ablation study on physics-informed information (iRadioDiff).}
\captionsetup{font={small}, skip=6pt}
\vspace{-2pt}
\resizebox{0.95\linewidth}{!}{
\begin{tabular}{@{}cc|ccccc@{}}
\toprule
\multicolumn{2}{c|}{Methods} & \textbf{RMSE} & \textbf{PSNR} & \textbf{LPIPS} & \textbf{FID} & \textbf{Loc. Error (m)} \\
\midrule
\multirow{2}{*}{ALG} & w/o Physics & 9.619 & 28.65 & 0.6260 & 351.1 & 8.553 \\
                     & w/ Physics & {\color[HTML]{9A0000} \textbf{6.357}} & {\color[HTML]{9A0000} \textbf{32.24}} & {\color[HTML]{9A0000} \textbf{0.2742}} & {\color[HTML]{9A0000} \textbf{145.2}} & {\color[HTML]{9A0000} \textbf{7.860}} \\
\midrule
\multirow{2}{*}{ZLG} & w/o Physics & 11.75 & 27.06 & 0.3717 & 242.1 & 8.765 \\
                     & w/ Physics & {\color[HTML]{9A0000} \textbf{7.010}} & {\color[HTML]{9A0000} \textbf{31.45}} & {\color[HTML]{9A0000} \textbf{0.3301}} & {\color[HTML]{9A0000} \textbf{192.6}} & {\color[HTML]{9A0000} \textbf{8.830}} \\
\bottomrule
\end{tabular}
}
\vspace{-8pt}
\label{tab3}
\end{table}
To evaluate the effectiveness of the proposed physics-informed strategy, we conduct ablation studies on iRadioDiff by removing all physics-related priors. In the ablated setting, the model receives only the raw environmental descriptors and AP location—consistent with baseline methods such as RadioDiff and RadioUNet—without incorporating reflection, transmission, or boundary-derived discontinuity information. As shown in Table~\ref{tab3}, the full physics-informed iRadioDiff consistently outperforms its ablated variant across all evaluation metrics, including both radio map construction quality and localization accuracy. These results clearly demonstrate the effectiveness of integrating physically grounded environmental cues into the generative process.

\section{Conclusion}
In this paper, we have presented iRadioDiff, a physics-informed diffusion framework for sampling-free indoor radio map construction. By encoding electromagnetic material properties, including reflection/transmission, geometry-derived diffraction and strong-transmission boundaries, and LoS contours, the approach injects physically consistent priors via cross-attention, enabling faithful recovery of nonstationary field structures that conventional data-driven models struggle to capture. Extensive experiments have shown SOTA performance in RM construction and improved RSSI-based localization, validating the efficacy of physics-guided conditioning. The method is well-suited to communication networks and positioning services that demand scalable, rapidly deployable, and environment-adaptive RM generation. Future work will extend the framework to multi-AP and 3D settings, incorporate dynamic scenario changes for real-time updates, and investigate joint training with downstream localizers for end-to-end optimization.

\bibliography{ref}
\bibliographystyle{IEEEtran}
\ifCLASSOPTIONcaptionsoff
  \newpage
\fi
\end{document}